\documentclass[runningheads]{llncs}

\usepackage[T1]{fontenc}
\usepackage{graphicx}
\usepackage{amsmath}
\usepackage{multirow}
\usepackage{booktabs}
\usepackage{url}
\usepackage{hyperref}
\begin{document}
\title{PIPsUS: Self-Supervised Point Tracking in Ultrasound\thanks{Supported by NSERC Discovery Grant and Charles Laszlo Chair in Biomedical Engineering held by Dr. Salcudean, VCHRI Innovation and Translational Research Awards, and the University of British Columbia Department of Surgery Seed Grant held by Dr. Prisman. This work was completed when Dr. Schmidt was at the University of British Columbia.}}
%
%
\author{Wanwen Chen\inst{1} \and
Adam Schmidt\inst{2}
\and Eitan Prisman\inst{3} \and
Septimiu E Salcudean\inst{1,4}}
%
\authorrunning{W. Chen et al.}

\institute{Department of Electrical and Computer Engineering, The University of British Columbia, Vancouver, BC, Canada \and
Intuitive Surgical Inc., Sunnyvale, CA, United States \and
Division of Otolaryngology, Department of Surgery, The University of British Columbia, Vancouver, BC, Canada \and
School of Biomedical Engineering, The University of British Columbia, Vancouver, BC, Canada\\
\email{\{wanwenc,tims\}@ece.ubc.ca}}


%
\maketitle              
%

\begin{abstract}
Finding point-level correspondences is a fundamental problem in ultrasound (US), enabling US landmark tracking for intraoperative image guidance and motion estimation. 
Most US tracking methods are based on optical flow or feature matching, initially designed for RGB images. Therefore domain shift can impact their performance.
Ground-truth correspondences could supervise training, but these are expensive to acquire.
To solve these problems, we propose a self-supervised point-tracking model called PIPsUS.
Our model can track an arbitrary number of points at pixel-level in one forward pass and exploits temporal information by considering multiple, instead of just consecutive, frames.
We developed a new self-supervised training strategy that utilizes a long-term point-tracking model trained for RGB images as a teacher to guide the model to learn realistic motions and use data augmentation to enforce tracking from US appearance. 
We evaluate our method on neck and oral US and echocardiography, showing higher point tracking accuracy when compared with fast normalized cross-correlation and tuned optical flow. 
Codes are available at \url{https://github.com/aliciachenw/PIPsUS}.

\end{abstract}
\section{Introduction}
Intraoperative ultrasound (US) in head and neck surgery is an emerging tool that helps surgeons localize tumors and important anatomy such as arteries.
Keypoint tracking can assist surgeons in finding and keeping tissue of interest in the US plane and estimating relative transducer motion~\cite{zhao2022uspoint}.
Moreover, finding point-level correspondences in US can benefit many clinical applications, such as large strain estimation and image registration, where deformation can be modeled by the motion of control points.
Sparse feature matching and optical flow are two different methods of finding point correspondences.
The former requires designing feature descriptors and is usually limited to keypoints that are detected by a detector, while the latter estimates dense pixel-level motion for consecutive frames instead of long-term motion.
Recent work in particle video~\cite{harley2022particle,zheng2023pointodyssey} predicts the motion of densely-sampled points based on feature correlation and pixel motion, exploits both sparse feature matching and optical flow and achieves high accuracy in long-term pixel tracking. 

Though point tracking has been studied in RGB images, its implementation in US is difficult.
Most models for RGB images are trained on labeled datasets~\cite{doersch2022tap} or simulated images~\cite{zheng2023pointodyssey}.
However, it is difficult to simulate 2D US videos because of the complexity of the physical interaction between sound waves and tissue. 
Labeling corresponding points is time-consuming and expensive, and requires training labelers to understand the particulars of US such as artifacts.

Targeting the problems above, we enable point tracking in US utilizing a particle video representation with a self-supervised training strategy.
Our novelty includes: 
(1) A new particle video model for online tracking of any number of points at the same time in US image sequences;
(2) A new self-supervised teacher-student learning strategy: a teacher model that can view entire clips is used to train our model to estimate frame-by-frame motion.
 
\section{Related Work}

\textbf{Feature and Template-based Matching in US:}
Feature matching has been used to track landmarks in US.
Early methods used detectors and descriptors for RGB images such as FAST~\cite{wulff2019robust} and SIFT~\cite{machado2018non}. 
However, these methods do not consider the special texture of US.
Some methods use handcrafted features for US~\cite{alkhatib2018adaptive,dall2015bipco} but do not achieve real-time performance.  
Recent success of deep learning-based keypoint matching in RGB images raises interest in their applications to US, such as in Zhao \textit{et al.}~\cite{zhao2022uspoint} and Shen \textit{et al.}~\cite{shen2019discriminative}. 
However, labels or external sensors are required for their supervised training.

Self-supervised or unsupervised learning has been investigated to reduce the effort of labeling. In~\cite{wulff2023landmark}, an autoencoder is trained to reconstruct patches and the latent space vector for each patch is used as feature for tracking.
Learning from data augmentation~\cite{wang2022multi3,liu2020cascaded} and contrastive learning with teacher-student learning~\cite{liang2023semi} have also been employed.
However, these methods have been tested on a limited number of landmarks (usually on the CLUST dataset which has 1-5 annotated landmarks for each sequence~\cite{de2018evaluation}).
Compared with these methods, our model can track an arbitrary number of points in one forward pass. 

\textbf{Optical Flow:}
Optical flow estimates dense pixel-level motion and is similar to our work, but it is not widely applied in US. 
Classical optical flow methods such as Lukas-Kanade~\cite{makhinya2015motion}, 
optimization-based flow~\cite{ouzir2018robust}, or in combination with block-matching~\cite{chuang2017tendon} have been used for US, and deep learning optical flow such as LiteFlowNet~\cite{al2022object}, FlowNet~\cite{evain2020pilot} and PDD-Net~\cite{nicke2022realtime} have also been investigated.
However, these models only consider consecutive frames, making them more sensitive to drift and presenting artifacts in long-term tracking.
Unlike optical flow, our method uses features and motion history to improve tracking performance.

\textbf{Track-any-point in RGB:}
Our work is inspired by tracking-any-point (TAP), estimating pixel correspondences in videos.
TAP models such as PIPs~\cite{harley2022particle}, TAP-Net~\cite{doersch2022tap}, and PIPs++~\cite{zheng2023pointodyssey} are usually trained with real labels or simulated scenes.
Wang \textit{et al.}~\cite{wang2023tracking} proposed a new TAP model that is supervised by optical flow. 
However, most established TAP models infer whole sequences, limiting the usability in intra-operative US tracking. We developed an online point-tracking model and a new learning strategy to train the model without manual labels.

\section{Methods}
\begin{figure}[tb]
\centering
\includegraphics[width=0.95\linewidth]{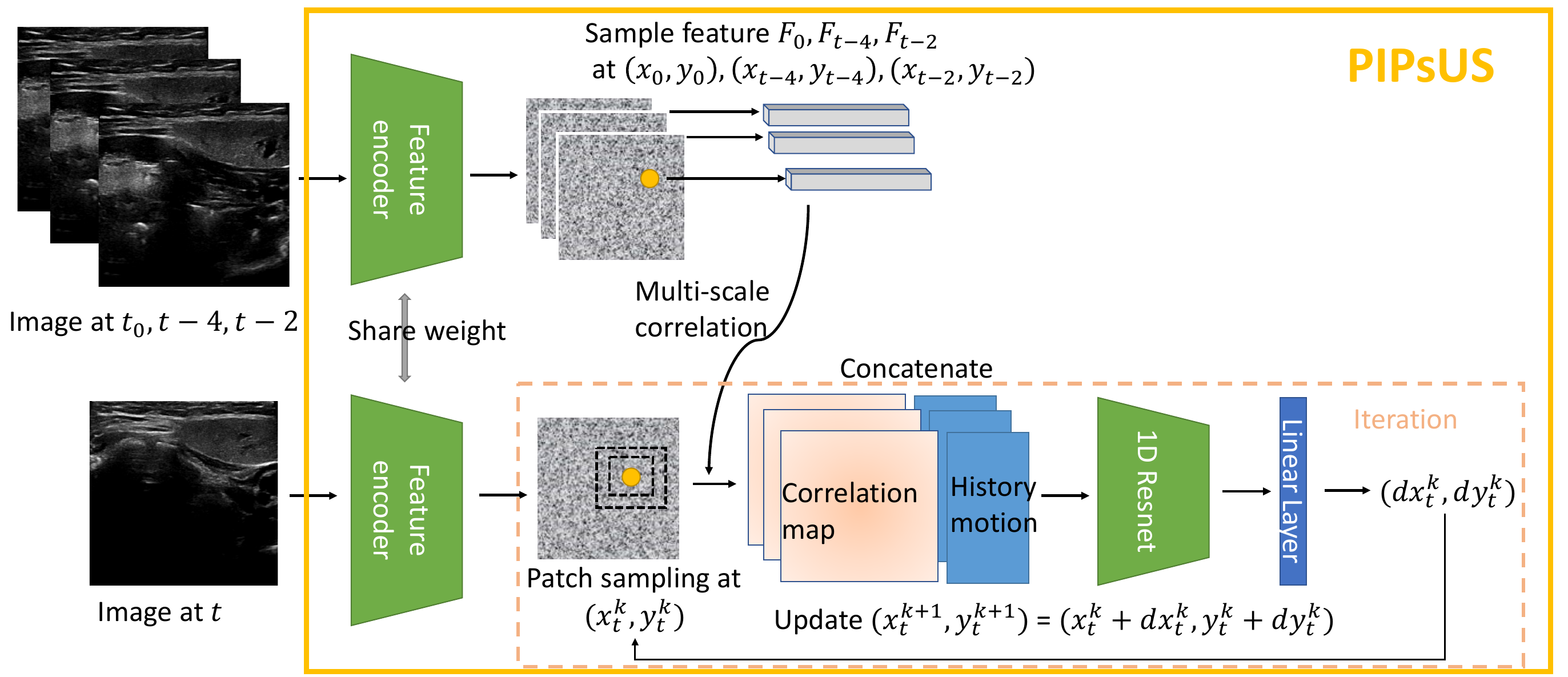}
\caption{PIPsUS architecture: PIPsUS enables streaming evaluation of point motion, estimating point motion at time $t$ using motion and image feature history. The model encodes history and current images and samples the features of the tracked points on history feature maps. The correlation maps of the history feature and current feature maps are concatenated with the history motion. A 1D-Resnet encodes the information and a linear layer iteratively predicts the tracking update.}\label{fig:model_structure}
\end{figure}
\textbf{PIPsUS Model.}
We propose a model named Persistent Independent Particles in US (PIPsUS) to track points in US, as shown in Fig.~\ref{fig:model_structure}. PIPsUS is inspired by PIPs++~\cite{zheng2023pointodyssey}, an improved model of Persistent Independent Particles (PIPs)~\cite{harley2022particle}, where pixel motions are modeled as moving particles in videos.
However, PIPs++ inspects the entire video to predict point trajectories, so it can not predict in a streaming manner, and the computation and memory cost increase regarding the length of the videos. The main advantage of our model PIPsUS is that it estimates motion in a streaming manner to keep computation and memory costs constant, allowing it to run intra-operatively with fewer memory resources. It first has a feature encoder that encodes the US images into feature maps.
We use the same ResNet-based encoder as PIPs++ to reuse the pre-trained weights to accelerate convergence.
Features of the tracked points are sampled at prior frames by $F_i=bilinearsampling(I_i, \mathbf{p}_i=(x_i,y_i))$, where $i\in \{0,t-4,t-2\}$.
Sampling from multiple frames allows the model to learn the original and current appearance of tracked points. 
The encoder also encodes the image $I_t$ to generate a dense feature map.
The new point location $\mathbf{p}_t=(x_t,y_t)$ is estimated using an iterative method that showed great success in RAFT~\cite{teed2020raft} and PIPs++.
We assume zero-motion to start, so the initial location $\mathbf{p}_t^0=(x_t^0,y_t^0)$ is $\mathbf{p}_{t-1}$.
For each iteration $k$, we sample an $R\times R$ patch in the current frames's feature map at $\mathbf{p}_t^k$ as $P^0,P^1,...,P^{L}$, where $L$ is the number of resolution layers.
Feature $F_i$ is correlated with these patches and the resulting correlation maps at each resolution are concatenated and reshaped into $n$ $L\times R^2$ vectors, where $n$ is the number of points. 
Since the point motion should be consistent, we concatenate the most recent motion flow $\mathbf{p}_t^k-\mathbf{p}_{t-1}, \mathbf{p}_t^k-\mathbf{p}_{t-2}, \mathbf{p}_t^k-\mathbf{p}_{t-3}$ and use a sinusoidal position embedding~\cite{harley2022particle} to generate motion vectors to enable the model viewing recent motion.
A 1D-Resnet then encodes the concatenated motion and correlation vectors, and a linear layer is used to predict the update $\Delta \mathbf{p}_t^k$.
We update $\mathbf{p}_t^{k+1}=\mathbf{p}_t^k+\Delta \mathbf{p}_t^k$ for the next iteration.
To start the inference when there is no enough previous frame, we pad the video with $I_0$ and $\mathbf{p}_0$ since our model needs image and location history.

\textbf{Self-supervised Teacher-Student Training:}
To train the model without manual labels, we use two different pseudo-ground truth labels:

(1) PIPs++ teacher labels: We use point trajectories predicted by PIPs++ as labels to guide PIPsUS to predict the point motions. 
We use a weighted Huber loss because it is more robust against outliers than L1 loss, preventing the model from overfitting possible wrong predictions in PIPs++:
\begin{equation}
L_t=\mu_t\sum_{k=0}^{K} w_k HuberLoss(\mathbf{p}_t^{gt}, \mathbf{p}_t^k)\label{eq:pips_loss}
\end{equation}
$w_k$ is an increased weight with update iteration $w_k=\gamma_{Iter}^{K-k-1}$ to encourage the model to learns the update function. The weight $\mu_t=\gamma_{Time}^{T-t-1}$ increases with time to encourage the model to reduce drifting. We set $\gamma_{Iter}=0.8$ and $\gamma_{Time}=0.95$.

(2) Simulation labels: We randomly transform ultrasound images with translation, intensity modulation, and noise addition to generate US videos with known motions.
The loss is similar to Eq.~\ref{eq:pips_loss} but with $\gamma_{Time}=1$, and we choose L1 loss since the motion under our transformations is known.

We use simulation to warm up the model, then train the model with PIPs++ labels plus 50\% of the simulated labels and validate on PIPs++ labels.
We implement a zero-flow regularization with zero-motion videos to regularize model prediction.
The input motion history has a 70\% chance of being the accurate flow with added Gaussian noise and 30\% to be the model's previous prediction.
As a proof-of-concept experiment, the keypoints are detected by SIFT, with the contrast threshold as 0.08 and the edge threshold as 4. 
However, our model is not limited to SIFT in training and inference.

\section{Experiments}

\textbf{Data:}
(1) Neck and oral US (OUS): This is a private dataset containing 2D US sequences collected from 19 patients who underwent transoral robotic surgery, from January 2022 to October 2023 at the Vancouver General Hospital (Vancouver, BC, Canada). This study received ethics approval from the UBC Clinical Research Ethics Board (H19-04025).
A BK3500 and a 14L3 linear 2D transducer (BK Medical, Burlington, MA) were used in the operation room for US imaging and a Polaris Spectra (Northern Digital, ON, Canada) was used to track the US transducer. 
PLUS~\cite{Lasso2014a} was used to record the US videos.
The image depth is 4~cm at 9 MHz, with a frame rate of $5.76\pm0.89$ fps.
For each patient, the US scan included the neck, oropharynx, and the base of tongue (BOT) on the cancerous side, before and after the tongue retraction. 
Data from 12 patients are used for training, 4 patients for validation, and 3 patients for testing. 
We did not include the BOT scan for PIPs++ ground truth, since it mainly contains out-of-plane motions along the neck, but we kept the BOT images in the simulated sequences.
For the PIPs++ labels, we split the recorded sequences into 20 frame-long sequences. 
For the simulated labels, we used the first frame in the videos to generate 41 frame-long sequences.
Images were resized to $256\times256$. The final amount of data is summarized in Table~\ref{tab:dataset}.

(2) EchoNet~\cite{ouyang2019echonet}: EchoNet is a public dataset containing 2D videos of US cardio motion, shared under the Stanford University Dataset Research Use Agreement.
We randomly select 200 videos for training, 50 videos for validation, and 50 videos for testing. Images were resized to $256\times256$.
We use the same method in OUS to generate US sequences and 
the dataset statistics is shown in Table~\ref{tab:echonet_dataset}.

\begin{table}[tb]
\centering
\caption{Number of collected videos and generated sequences in OUS dataset.}\label{tab:dataset}
\begin{tabular}{cccccc}
\hline
                           &            & Videos & Frames & \# generated sequences & \# keypoints per sequence   \\
\hline
\multirow{3}{*}{Pips++}     & Train   & 174          & $65\pm42$        & 452 & $18.75\pm14.33$ \\
                           & Valid & 57           & $112\pm70$      & 278 & $14.50\pm11.20$ \\
                           & Test       & 45           & $116\pm82$       & 225 & $8.88\pm8.19$   \\
\hline
\multirow{2}{*}{Sim}  & Train   & 212          & 41           & 199 & $20.53\pm13.78$ \\
                           & Test       & 59           & 41           & 49  & $13.00\pm9.57$ \\
\hline
\end{tabular}
\end{table}

\begin{table}[tb]
\centering
\caption{Number of videos and generated sequences used in EchoNet dataset.}\label{tab:echonet_dataset}
\begin{tabular}{cccccc}
\hline
                           &         & Videos & Frames      & \# generated sequences & \# keypoints per sequence   \\
\hline
\multirow{3}{*}{Pips++}    & Train   & 200    & $173\pm46$    & 1620                   & $69.88\pm19.97$ \\
                           & Valid   & 50     & $168\pm38$  & 392                    & $68.65\pm17.58$ \\
                           & Test    & 50     & $175\pm47$  & 410                    & $75.87\pm21.33$   \\
\hline
\multirow{2}{*}{Sim}       & Train   & 200    & 41          & 200                    & $75.36\pm20.41$ \\
                           & Test    & 50     & 41          & 50                     & $81.72\pm24.36$ \\
\hline
\end{tabular}
\end{table}

\textbf{Training Details:}
The models were trained on an NVIDIA Tesla V100 and implemented in python 3.8, PyTorch-2.1.0, and CUDA-11.8.
The weight of the encoder was initialized with the public weight of PIPs++.
AdamW optimizer was used with a learning rate of 5e-4 for warmup, and 1e-4 for self-supervised teacher-student tuning.
We warmed up the model with 10 epochs and then the model was trained with our self-supervised learning method for 50 epochs.
The models with the lowest loss on the validation set were selected.
The image batch size was 1 because the number of keypoints per batch varies.

\begin{table}[tb]
\centering
\caption{Quantitative evaluation on OUS. The L2 error is in pixels. ** The average time for PIPs++ to run inference on each sequence (20 frames) is 0.08 seconds.}\label{tab:quant_sift_results} 
\begin{tabular}{c|cc|ccc}
\hline
           & \multicolumn{2}{|c|}{Simulation}          & \multicolumn{3}{c}{Real Data}         \\
\hline
Method     & L2                 & NCC                 & L2                   & NCC       & FPS    \\
\hline
Fast NCC   & 6.65±6.25         & 0.83±0.19           & 22.84±27.91          & 0.82±0.19  & 145.2\\
RAFT       & 21.49±21.20        & 0.80±0.21           & 14.65±14.57          & 0.80±0.21  & 60.2 \\
PIPs++     & \textbf{0.94±0.72} & \textbf{0.95±0.10}  & N/A                  & N/A              & N/A**\\
PIPsUScorr & 1.28±1.19          & 0.94±0.12           & 12.10±16.95          & \textbf{0.84±0.19}  & 43.3 \\
PIPsUS     & 1.10±0.88          & \textbf{0.95±0.11}  & \textbf{9.04±11.79}  & \textbf{0.84±0.19}   & 34.6\\
\hline
\end{tabular}
\end{table}

\begin{table}[tb]
\centering
\caption{Quantitative evaluation on EchoNet. The L2 error is in pixels. ** The average time for PIPs++ to run inference on each sequence (20 frames) is 0.13 seconds.}\label{tab:quant_sift_results_echo} 
\begin{tabular}{c|cc|ccc}
\hline
           & \multicolumn{2}{c|}{Simulation}             & \multicolumn{3}{c}{Real Data}         \\
\hline
Method     & L2                    & NCC                 & L2                 & NCC                  & FPS    \\
\hline
Fast NCC   & 6.40±6.60             & 0.83±0.22           & 5.05±10.56         & 0.93±0.13            &  19.3\\
RAFT       & 12.27±11.25           & 0.80±0.24           & 4.32±3.60          & 0.86±0.20            &   70.0\\
PIPs++     & \textbf{0.90±0.75}             & \textbf{0.96±0.11}           & N/A                & N/A                  &   N/A**\\
PIPsUScorr & 1.26±1.21    &  0.95±0.12  & \textbf{2.27±2.62} & \textbf{0.94±0.11}   &   39.0\\
PIPsUS     & 1.07±1.04             & 0.95±0.11           & 2.40±3.15          & \textbf{0.94±0.11}            &  15.4\\
\hline
\end{tabular}
\end{table}

\section{Results}
We compare our models with fast normalized cross-correlation (NCC) template matching~\cite{lewis2001fast}, fine-tuned RAFT~\cite{teed2020raft}, PIPs++, and conduct an ablation study. PIPsUS has correlation maps and motion history in the input, while PIPsUScorr only sees the correlation. RAFT was fine-tuned with US using mean square loss between target images and optical flow-wrapped source images for 5 epochs. We choose RAFT because it outperforms LiteFlowNet and FlowNet~\cite{teed2020raft}, and PDD-Net~\cite{nicke2022realtime} is originally designed for 3D registration.
We evaluated the performance on simulated US sequences with smooth random affine motions and intensity changes and real US sequences.
Due to the lack of ground truth labels on real US, we compare with teacher model PIPs++ pseudo ground truth, similar to~\cite{ihler2020self}.
The quantitative evaluation includes L2 error and image-patch NCC, as shown in Table~\ref{tab:quant_sift_results} for OUS and Table~\ref{tab:quant_sift_results_echo} for EchoNet. The L2 error evaluates the absolute tracking accuracy, while the NCC quantifies the keypoint patch similarity. 

\begin{figure}[tbp]
\centering
\includegraphics[width=0.49\linewidth]{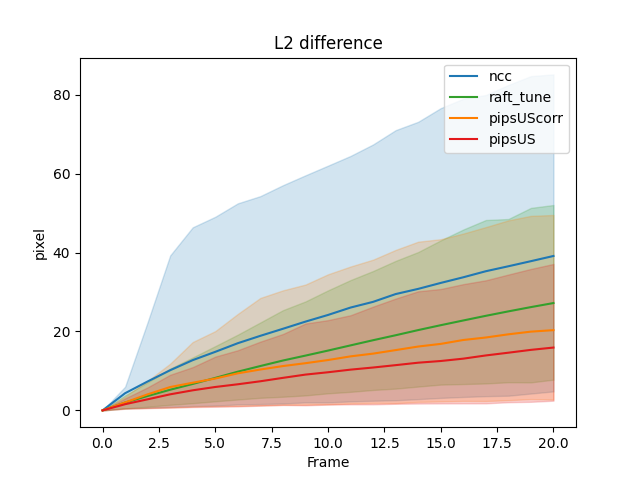}
\includegraphics[width=0.49\linewidth]{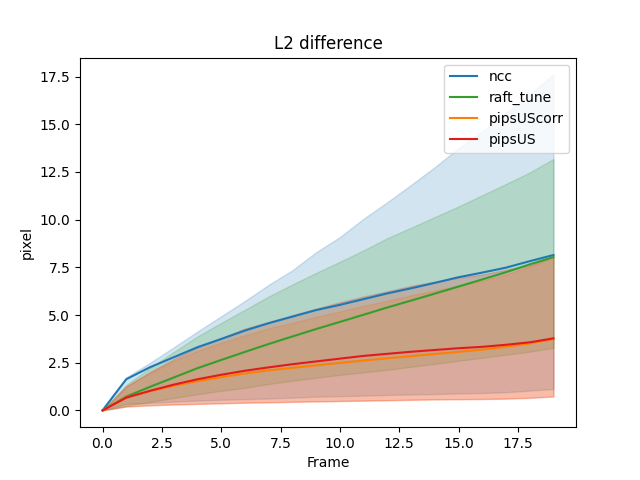}
\caption{L2 in different frames on real US sequence. Left: on OUS, right: on EchoNet. The line is average L2 and the shadow is 10 and 90 percentile.}\label{fig:test_timeseries_results}
\end{figure}
\begin{figure}[tb]
\centering
\includegraphics[width=0.95\linewidth]{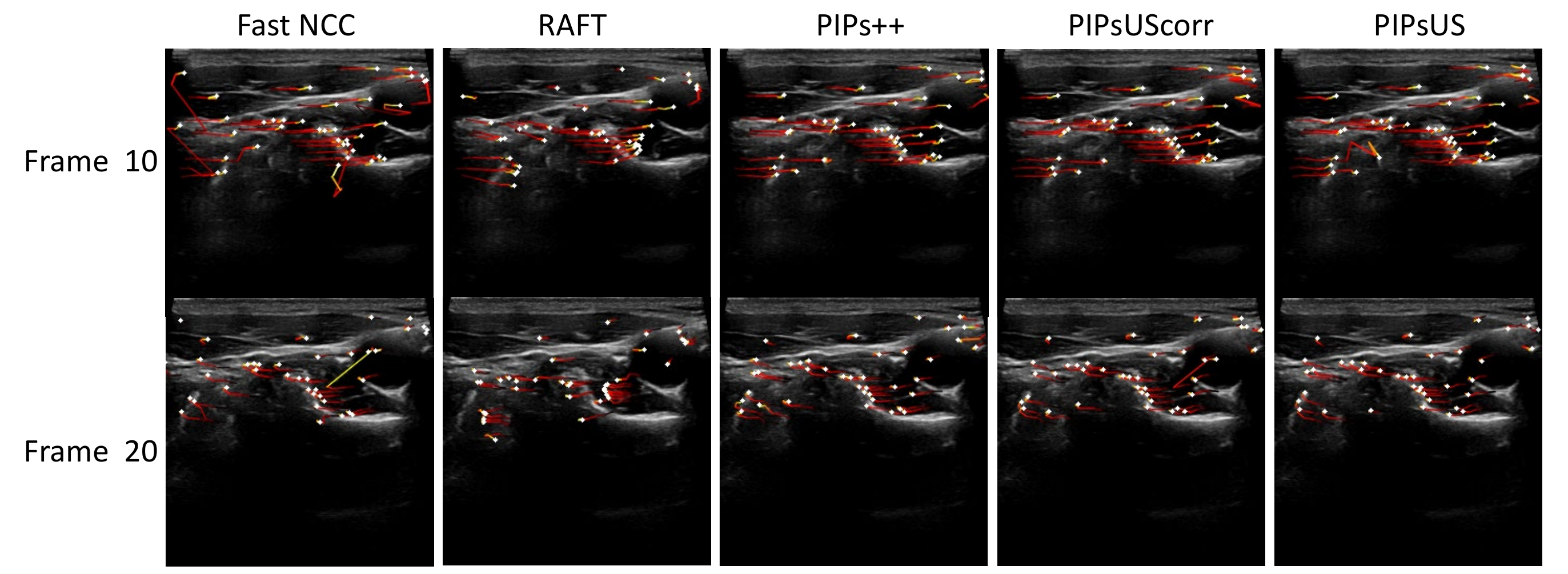}
\includegraphics[width=0.95\linewidth]{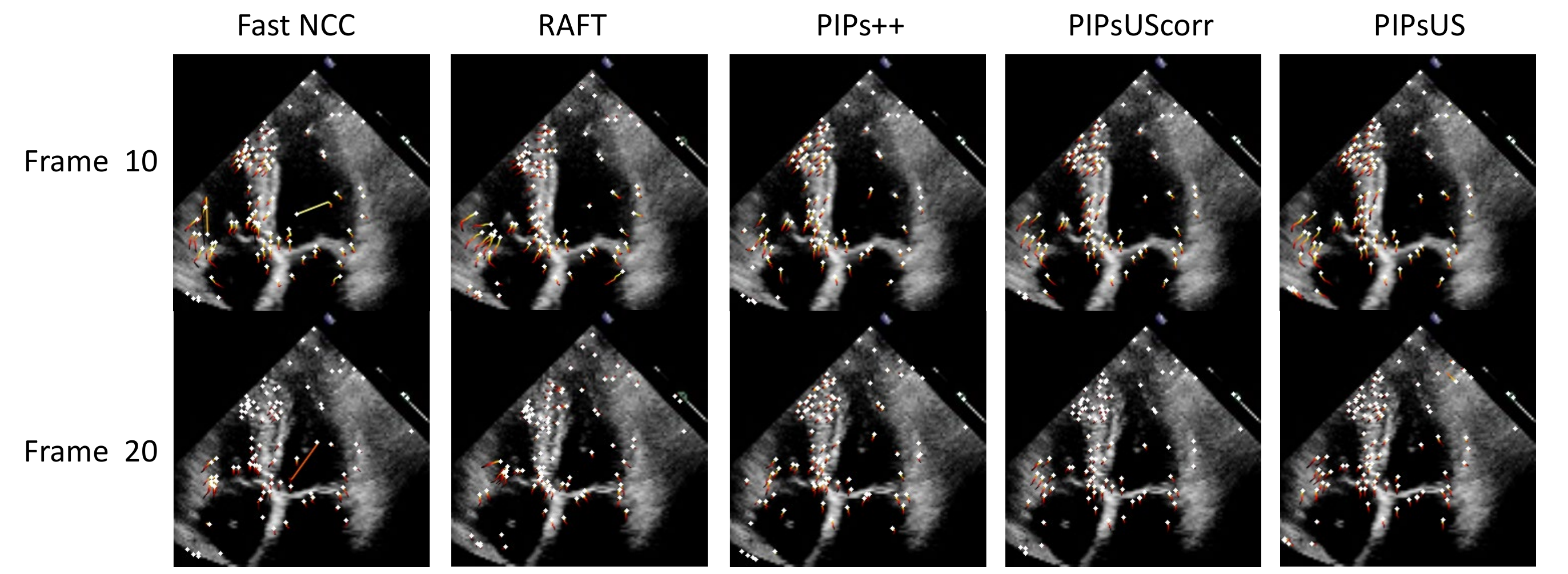}
\caption{Examples of tracked point trajectories in different frames on OUS (top 2 rows) and EchoNet (bottom 2 rows). The point is the current predicted keypoint locations and the colored line is the trajectory history. On OUS, in Frame 20 of PIPsUScorr and NCC, a point is correlated to a faraway location. By using point motion history, PIPsUS avoids this.}\label{fig:example}
\end{figure}

In the simulation, PIPs++ has the highest accuracy and patch similarity, but PIPsUS is comparable to PIPs++ both on the OUS and EchoNet. PIPsUS and PIPsUScorr all demonstrate a large improvement in accuracy and similarity compared with fast-NCC and RAFT. However, PIPs++ needs to infer the whole video sequences, while PIPsUS can infer the point location in a frame-by-frame manner, making PIPsUS preferable in online applications such as intra-operative landmark tracking.
In the real US, PIPsUS achieves the highest accuracy on OUS. PIPsUScorr performs the best on EchoNet, but PIPsUS is comparable. Again, PIPsUS and PIPsUScorr achieve higher accuracy and patch similarity in both datasets compared with Fast-NCC and RAFT. All methods perform better on EchoNet than OUS, and we expect this is because echocardiography contains smaller in-plane motions from standard views while freehand OUS contains larger motions and points can be out-of-plane. 
Thus, tracking points in EchoNet is easier.
Fig.~\ref{fig:test_timeseries_results} shows the trend of L2 error; fast NCC and RAFT are both sensitive to drift. 
PIPs-like models view the features at different times, so their performance degrades slowly. 
The results show the advantages of investigating multiple frames instead of just consecutive frames. 
The comparison between PIPsUS and PIPsUScorr shows that the motion history can improve motion estimation in OUS but not in EchoNet. 
We hypothesize that motion history can help generate a reasonable prediction when points have larger motion and are temporally out-of-plane, so its advantage is not shown in echocardiography where features remain in-plane. 
This is also shown in Fig.~\ref{fig:example} which displays images of the tracked points and trajectories in OUS and EchoNet. More visualization is provided in the supplemental materials.
We also evaluate the survival rate, which is the percentage of points that have an L2 error smaller than 50, as defined in~\cite{zheng2023pointodyssey}. 
In real EchoNet, the survival rate of the points at the end of sequences is 97.27\% for fast NCC, and 100.00\% for other methods. For OUS, the survival rate is 67.95\% for fast NCC, 88.89\% for RAFT, 90.27\% for PIPsUScorr and 95.19\% for PIPsUS, showing that PIPsUS is better at keeping track of points in  OUS, which has more complicated motions.

\textbf{Limitations and Future Work:} 
Though we show improved tracking, disappearing and reoccurring detection is required for longer-term landmark tracking, which will be the next step in our investigation. 
We did not investigate other data augmentation methods, and US physics-based augmentation might improve the model's robustness against deformation, shadows, and artifacts.
Our model relies on feature correlation and temporal information without the awareness of point saliency. 
Integrating segmentation maps might allow the model to focus on tracking anatomically salient landmarks. 
Also, SIFT may not detect all anatomically meaningful keypoints, and we expect that new detectors can be designed to find more suitable keypoints for tracking. 
These limitations have not been addressed in previous US point tracking either. 


\section{Conclusions}
We propose a new model utilizing particle video to track an arbitrary number of points utilizing dense feature maps and particles' previous motion with constant memory cost, and can track points in an on-line manner.
We develop a self-supervised teacher-student training strategy to train our model to learn without manual labeling.
Our model achieves higher accuracy compared with fast NCC and fine-tuned RAFT, and it is more robust to temporal drift.

\bibliographystyle{splncs04} 
\bibliography{reference} 
\end{document}